\pdfoutput=1
\documentclass[11pt]{article}

\usepackage[]{acl}

\usepackage{times}
\usepackage{latexsym}
\usepackage{booktabs}
\usepackage[T1]{fontenc}
\usepackage[utf8]{inputenc}

\usepackage{microtype}

\usepackage{hyperref}
\usepackage{graphicx}
\graphicspath{{images/}}
\usepackage{multirow}
\usepackage{enumitem}
\usepackage{todonotes}

\usepackage{amsmath}
\usepackage{comment}

\title{ Explaining Translationese: why are Neural Classifiers Better\\ and what do they Learn? }

\author{Kwabena Amponsah-Kaakyire\textsuperscript{{\normalfont *1,2}}, Daria Pylypenko\textsuperscript{{\normalfont *1}}, {\bf Josef van Genabith}\textsuperscript{{\normalfont 1,2}}{\bf,} \\ \and {\bf Cristina Espa\~{n}a-Bonet\textsuperscript{{\normalfont 2}}} \\  
  \textsuperscript{1}Saarland University, \textsuperscript{2}German Research Center for Artificial Intelligence (DFKI) \\
  Saarland Informatics Campus, Saarbrücken, Germany
  \\
  {\tt amponsahkaakyirek@gmail.com}\\
  {\tt daria.pylypenko@uni-saarland.de}\\
  {\tt \{cristinae, Josef.Van\_Genabith\}@dfki.de }\\
  } 

\begin{document}
\maketitle
\begin{abstract}
\renewcommand{\thefootnote}{\fnsymbol{footnote}}
\footnotetext[1]{Equal contribution.}
Recent work has shown that neural feature- and representation-learning, e.g. BERT, achieves superior performance over traditional manual feature engineering based approaches, with e.g. SVMs, in translationese classification tasks. Previous research did not show $(i)$ whether the difference is because of the features, the classifiers or both, and $(ii)$ what the neural classifiers actually learn.  To address $(i)$, we carefully design experiments that swap features between BERT- and SVM-based classifiers. We show that an SVM fed with BERT representations performs at the level of the best BERT classifiers, while BERT learning and using handcrafted features performs at the level of an SVM using handcrafted features. This shows that the performance differences are due to the features. To address $(ii)$ we use integrated gradients and find that $(a)$ there is indication that information captured by hand-crafted features is only a subset of what BERT learns, and $(b)$ part of BERT's top performance results are due to BERT learning topic differences and spurious correlations with translationese.

\end{abstract}

\section{Introduction}
\label{s:intro}

Translationese is a descriptive (non-negative) cover term for the systematic differences between translated and originally authored text in same language~\cite{gellerstam:1986}. Some aspects of translationese such as source interference \cite{toury:1980,Teich2003} are language dependent, others are presumed universal, e.g. simplification, explicitation, overadherence to target language linguistic norms \cite{volanskyEtAl:2015} in the products of translations.
While translationese effects can be subtle, 
especially for professional human translation, corpus-based studies \cite{Baker1993CorpusLA} and, in particular, machine-learning and classifier based studies \cite{rabinovich-wintner-2015-unsupervised,volanskyEtAl:2015,rubino-etal-2016-information,pylypenko-etal-2021-comparing} clearly reveal the differences.

While research on translationese is important from a theoretical point of view (translation universals, specific interference), it has a direct impact on machine translation research: \cite{kurokawa:2009, stymne2017effect, toral-etal-2018-attaining, zhangToral:2019, freitagEtAl:2019, graham-etal-2020-statistical, riley-etal-2020-translationese}, amongst others, show that translation direction in training and test data impacts on results, that already translated test data are easier to translate than original data, that  machine translation and post-editing result in translationese, and that mitigating translationese in MT output can improve results. Translationese impacts cross-lingual applications, e.g. question answering and natural language inference \citep{Singh2019XLDACD, clark-etal-2020-tydi, artetxe-etal-2020-translation}.

In this paper we focus on machine-learning-classifier-based research on translationese. Here, typically a
classifier is trained to distinguish between original and translated
texts (in the same language). Until recently, most of this research
\cite{Baroni2006ANA,volanskyEtAl:2015,rubino-etal-2016-information} used manually defined, often linguistically inspired,
feature-engineering based sets of features, mostly using support vector machines (SVM). 
Once a classifier is trained, feature importance and ranking methods are used to reason back to what aspects of the input is responsible for (i.e. explains) the classification (and whether this accords with linguistic theorisation). More recently, a small number of papers explored feature- and representation-learning neural network based approaches to translationese classification \cite{sominsky-wintner-2019-automatic}. In a
systematic study \citet{pylypenko-etal-2021-comparing} show that feature- and
representation-learning deep neural network-based approaches (in
particular BERT-based, but also other neural approaches) to
translationese classification substantially outperform handcrafted
feature-engineering based approaches using SVMs. However, to date, two important questions remain: $(i)$ it is not clear whether the substantial performance differences are due to learned vs. handcrafted features, the classifiers (SVM, the BERT classification head, or full BERT), or the combination of both, and $(ii)$ what the neural feature and representation learning approaches actually learn and how that explains the superior classification. The contributions of our paper are as follows:

\begin{enumerate}
    \item we address $(i)$ by carefully crossing features and classifiers, feeding BERT-based learned features to feature-engineering models (SVMs), feeding the BERT classification head with hand-crafted features, and by making BERT architectures learn handcrafted features, as well as feeding embeddings of handcrafted features into BERT. 
    Our experiments show that SVMs using BERT-learned features perform on a par with our best BERT-translationese classifiers, while BERT using handcrafted features only performs at the level of feature-engineering-based classifiers. This shows that it is the features and not the classifiers, that lead to the substantial (up to 20\% points accuracy absolute) difference in performance. 
    
    \item we present the first steps to address $(ii)$ using integrated gradients, an attribution-based approach, on the BERT models trained in various settings. Based on striking similarities in attributions between BERT trained from scratch and BERT pretrained on handcrafted features and fine-tuned on text data, as well as comparable classification accuracies, we find evidence that the hand-crafted features do not bring any additional information over the set learnt by BERT. it is therefore likely that the hand-crafted features are a (possibly partial) subset of the features learnt by BERT. Inspecting the most attributed tokens,
    we present evidence of 'Clever Hans' behaviour: at least part of the high classification accuracy of BERT is due to names of places and countries, suggesting that part of the classification is topic- and not translationese-based. Moreover, some top features suggest that there may be some punctuation-based spurious correlation in the data.
\end{enumerate}

\section{Related Work}
\label{s:sota}

\paragraph{Combining learned and hand-crafted features.}
\citep{kaas-etal-2020-team,prakash-tayyar-madabushi-2020-incorporating, lim_madabushi:2020} combine BERT-based and manual features in order to improve accuracy.
\citep{kazameini_etal:2020, ray-garain:2020, zhang-yamana-2020-wuy} concatenate BERT pooled output embeddings with handcrafted feature vectors for classification, often using an SVM, where the handcrafted feature vector might be further encoded by a neural network or used as it is. 
Our work differs in that we do not combine features from both models but swap them in order to decide whether it is the features, the classifiers or the combination that explains the performance difference between neural and feature engineering based models.
Additionally, our approach allows us to examine whether or not representation learning learns features similar to hand-crafted features.

\paragraph{Explainability for the feature-engineering approach to translationese classification.} 
To date, explainability in translationese research has mainly focused on quantifying handcrafted feature importance. Techniques include inspecting SVM feature weights \citep{avner:2016, pylypenko-etal-2021-comparing}, correlation \citep{rubino-etal-2016-information}, information gain \citep{ilisei:2010}, chi-square \citep{ilisei:2010}, decision trees or random forests \citep{rubino-etal-2016-information, ilisei:2010}, ablating features and observing the change in accuracy \citep{Baroni2006ANA, ilisei:2010}, training separate classifiers on each individual feature (or feature set) and comparing accuracies \citep{volanskyEtAl:2015, avner:2016}. For $n$-grams, the difference in frequencies between the original and translationese classes \citep{koppel-ordan:2011, van-halteren:2008}, and the contribution to the symmetrized Kullback-Leibler Divergence between the classes \citep{kurokawa:2009} have been used.

\paragraph{Explainability for the neural approach to translationese classification.}
To date, explainability methods for neural networks have not been widely explored. \citet{pylypenko-etal-2021-comparing} quantify to which extent handcrafted features can explain the variance in the predictions of neural models, such as BERT, LSTMs, and a simplified Transformer, by training per-feature linear regression models to output the predicted probabilities of the neural models and computing the $R^2$ measure. They find that most of the top features are either POS-perplexity-based, or bag-of-POS features. However, their method treats the neural network as a black-box, whereas we use a method that accesses the internals of the model.

\paragraph{Integrated Gradients (IG).}
In our work we use the Integrated Gradients method \citep{sundararajan:2017} for explainability. This method provides attribution scores for the input with respect to a certain class. IG calculates the integral of gradients of the model $F$ with respect to the input $x$ (token embedding), along the path from a baseline $x'$ (in our case, PAD token embedding) to the input $x$:
\begin{equation}
\begin{split}
    \texttt{IntegratedGrads}_i(x)::= (x_i - x'_i) \times \\ 
    \int_{\alpha=0}^{1} \frac{\partial F(x' + \alpha \times (x-x'))}{\partial x_i} d \alpha
\end{split}
\end{equation}

\smallskip
The strength of the Integrated Gradients method is that it satisfies two fundamental axioms (Sensitivity and Implementation Invariance), while many other popular attribution methods, like Gradients \citep{Simonyan14a}, DeepLift \citep{shrikumar:2017} and LRP \citep{lrp:2015} violate one or both of them. IG also satisfies the completeness axiom, that is, IG is comprehensive in accounting for attributions and does not just to pick the top label \cite{sundararajan:2017}.

\section{Experimental Settings}
\subsection{Data}
\label{ss:data}
For our experiments, we use the monolingual German dataset in the  \href{https://doi.org/10.5281/zenodo.5596238}{Multilingual Parallel Direct Europarl} (MPDE) \cite{amponsah-kaakyire-etal-2021-rely} corpus. The set contains 42k paragraphs with half of the texts German originals and the other half translations into German from Spanish (see statistics in Appendix \ref{ssA:data}). We perform paragraph-level classification with an average length of 80 tokens per training sample.

We additionally use an in-domain Europarl-based heldout corpus of around 30k paragraphs for training language models and $n$-gram quartile distributions on it. This corpus consists of original German texts only.

\subsection{Base Setup}
\label{ss:basics}
We compare the traditional SVM-based feature engineering approach, which has demonstrated high performance in previous translationese research, to the BERT model known to be very successful for various NLP tasks, including classification.
As base setup, we reproduce the models from \citet{pylypenko-etal-2021-comparing} for the two architectures and a new baseline: 
\begin{enumerate}
\itemsep0.2em 
\item \label{itemSVM} a linear \textbf{SVM} on 108-dimensional \textbf{handcrafted feature} vectors (with surface, lexical, unigram bag-of-PoS, language modelling and $n$-gram frequency distribution features\footnote{See \citep{pylypenko-etal-2021-comparing} for the detailed list of features.}). [\textbf{handcr.-features+SVM}]
\item \label{itemLinClf} a \textbf{linear classifier} (BERT classification head, simple linear FFN, except for difference in input dimension) trained on the 108-dimensional \textbf{handcrafted feature} vectors. [\textbf{handcr.-features+LinearClassifier}]
\item \label{itemBERT} off-the-shelf Google's \textbf{pretrained BERT}-base model (12 layers, 768 hidden dimensions, 12 attention heads) which we \textbf{fine-tune} on the MPDE corpus for translationese classification. [\textbf{pretrained-BERT-ft}]
\item \label{itemScratch} a BERT-base model with the same settings trained \textbf{from scratch} on MPDE for translationese classification. [\textbf{fromScratch-BERT}]
\end{enumerate}

For \ref{itemSVM}, we estimate $n$-gram language models with SRILM \cite{Stolcke:2002} and do POS-tagging with SpaCy.\footnote{\url{https://spacy.io/}}
For \ref{itemBERT}, we use multilingual BERT~\cite{devlinEtAl:2019} (BERT-base-multilingual-uncased),  and fine-tune with the \textit{simpletransformers}\footnote{\url{github.com/ThilinaRajapakse/simpletransformers}} library. We use a batch size of 32, learning rate of $4 \cdot 10^{-5}$, and  the Adam optimiser with epsilon $1 \cdot 10^{-8}$.

\paragraph{} 
To ensure fair and comprehensive treatment, we carefully explore many experiments and variations below: we exchange input features between BERT and SVM architectures by $(i)$ feeding BERT-learned features into SVMs (Section~\ref{ss:BERTintoSVM}), hand-crafted features into the BERT classification head, and $(ii$-$a)$ letting the full BERT architecture learn handcrafted feature vectors used by SVMs and $(ii$-$b)$ feeding handcrafted feature vectors as embeddings into the BERT model (Section~\ref{ss:SVMintoBERT}).

\subsection{SVM Classifier with BERT Features}
\label{ss:BERTintoSVM}

We train an SVM with linear kernel on the features learnt by the pretrained BERT model fine-tuned on the translationese classification task.
We use the output of the BERT pooler, which selects the last layer $[CLS]$ token vector, with linear projection and $tanh$ activation as our feature vector. We use:
\begin{enumerate}
\itemsep0.2em 
\item BERT's 768-dim pooled vector output, [\textbf{pretrained-BERT-ft+SVM}]
\item a 108-dim PCA projection of this vector. [\textbf{pretrained-BERT-ft+PCA$_{108}$+SVM}]
\end{enumerate}
The PCA projection allows us to match the handcrafted feature vector dimensionality.

\subsection{BERT with Handcrafted Features}
\label{ss:SVMintoBERT}
Apart from feeding hand-crafted feature vectors into a suitably adjusted BERT classification head [\textbf{handcr.-features+LinearClassifier}], we carefully design two strategies to force the full BERT architecture use the handcrafted features.

\paragraph{Pretraining on handcrafted feature prediction.}
First, we train a BERT-base model from scratch on the MPDE dataset to predict the handcrafted features. This regression model [\textbf{BERT-reg-full}] takes unmasked text as input and predicts continuous values (the 108 dimension vectors representing handcrafted features originally used in training the SVM). The complete feature vector is predicted at once, and the pretraining is done by minimizing MSE loss between the predicted and the ground truth vector.
The weights of this model encode the information of the handcrafted features. With this pretrained model,
\begin{enumerate}
\itemsep0.2em 
    \item we freeze the weights, replace the regression head (linear layer predicting 108 features) with a linear classifier (a BERT classification head predicting the original or translationese label) and train the classifier on the MPDE data for translationese classification, [\textbf{BERT-r2c-full-frozen]\footnote{\textbf{r2c} -- regression-to-classification}}
    \item we do not freeze but fine-tune on MPDE for the translationese classification task. [\textbf{BERT-r2c-full-ft}]
\end{enumerate}

The comparison between frozen and unfrozen weights is designed to provide us insights on the importance of representation learning in BERT.

We reproduce the same approach as above with a smaller BERT model with only 6 layers instead of 12 [\textbf{BERT-reg-half}]. Interestingly, according to the losses when training for predicting the handcrafted features, the smaller BERT-reg-half performs comparably to BERT-reg-full (0.0041136 vs 0.0041148 MSE). 
We then load the weights of the small 6 layer model into the embedding layer and the first 6 layers of a 12 layer non-pretrained BERT-base model and, similarly as before:
\begin{enumerate}[resume]
\itemsep0.2em 
    \item we freeze the loaded weights in the first 6 layers and train the remaining 6 layers and classifier on the translationese classification task, [\textbf{BERT-r2c-half-frozen}]
    \item we do not freeze but fine-tune on the translationese classification task with randomly-initialised weights for the other 6 layers. [\textbf{BERT-r2c-half-ft}]
\end{enumerate}

\begin{figure}[!t]
    \centering
    \includegraphics[width=0.45\textwidth]{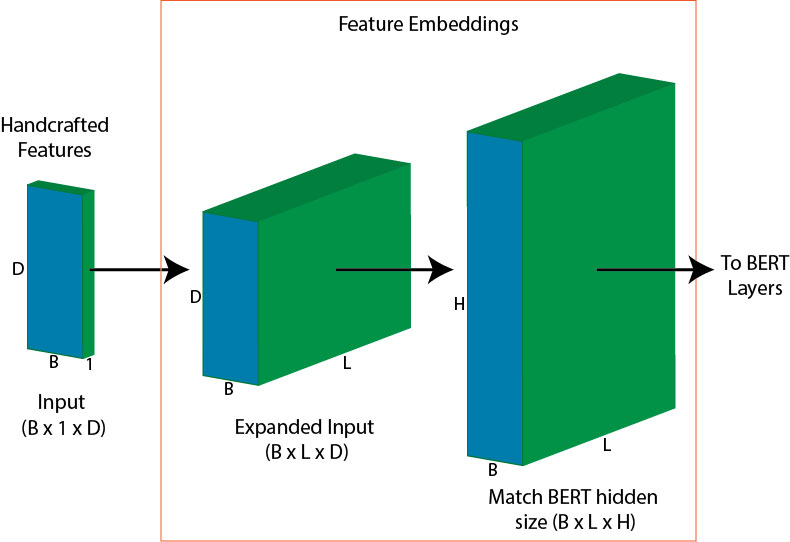}
    \caption{Mapping handcrafted features to embeddings.}
    \label{fig:bert-f2c}
\end{figure}

 \paragraph{Mapping handcrafted features to embeddings.}
 \label{model:f2c}
 Even though the very low MSE results indicate that both versions of BERT-reg are able to learn handcrafted features well, using them in terms of frozen layers in translationese classification leads to low classification performance (Section \ref{section:translationese_classification}). This could be attributed to the fact that, not being an end-to-end approach, information losses accumulate:
 first, even though MSE is low in BERT-reg, we do not have exactly the same features; and second, the features are not used directly for classification, but are encoded again by the network. This motivates us to explore an alternative way of encoding handcrafted features in an end-to-end manner.

 We convert the single vector of handcrafted features of dimension $D$ (108 in our experiments) into a sequence of embeddings in BERT's layer format, that is, length of feature embedding sequence $L$ times the dimension of the hidden states $H$ (768), while preserving the information of the single vector (Figure \ref{fig:bert-f2c}).
To do this, we consider a batch of tokens with size $B$ and take in the handcrafted features as a ($B \times D$)-dimensional input to the BERT model and generate feature embeddings by passing the features through 2 linear layers as follows. We first pass the ($B \times 1 \times D$) input to the first linear layer. The resulting ($B \times L \times D$)-dimensional output is fed as input to the second linear layer which outputs a ($B\times L\times H$)-dimensional output as the feature embeddings. 
 
This reshaped handcrafted feature embedding layer replaces BERT's embedding layer.
Weights are randomly initialised and the modified BERT model is trained on the translationese classification task. We experiment with three different values for sequence length $L$: 1, 80 (average length of our training samples) and 256 (half of maximum input for BERT). All three variants are trained from scratch [\textbf{BERT-f2c\footnote{\textbf{f2c} -- feature-embeddings to classification} L=1}, \textbf{BERT-f2c L=80}, \textbf{BERT-f2c L=256}]. For further comparison, we also take BERT-f2c L=80, load the weights of pretrained BERT-base layers into the 12 layers of the modified model and fine-tune on the task \textbf{[pretrained BERT-f2c L=80}].

Training and hyperparameter settings for these models are given in Appendix \ref{ssA:bert}.

\begin{table}[!t]
    \centering
    \small
    \begin{tabular}{|p{5cm}|c|}
    \hline\\
        [-0.8em]
        Model & Accuracy (\%) \\ \hline \\
        [-0.6em]
        handcr.-features+SVM & 73.2$\pm$0.1 \\
        handcr.-features+LinearClassifier & 72.0$\pm$0.4 \\
        pretrained-BERT-ft & 92.2$\pm$0.2 \\
        fromScratch-BERT & 89.3$\pm$0.3 \\
\midrule
        pretrained-BERT-ft+SVM & 92.0$\pm$0.0 \\
        pretrained-BERT-ft+PCA$_{108}$+SVM & 92.0$\pm$0.0 \\
\midrule
        BERT-r2c-full-frozen+SVM & 74.9$\pm$0.7 \\
        BERT-r2c-full-frozen+PCA$_{108}$+SVM & 70.3$\pm$0.1 \\
        BERT-r2c-full-frozen & 59.6$\pm$0.1 \\
        BERT-r2c-full-ft & 89.3$\pm$0.4 \\
        BERT-r2c-half-frozen & 67.5$\pm$0.4 \\
        BERT-r2c-half-ft & 89.0$\pm$0.3 \\
\midrule
        BERT-f2c $L = 1$ & 57$\pm$10 \\
        BERT-f2c $L = 80$ & 72.8$\pm$0.2 \\
        BERT-f2c $L = 256$ & 72.7$\pm$0.2 \\
        pretrained-BERT-f2c $L = 80$ & 68.0$\pm$2.1 \\ \hline
    \end{tabular}
    \caption{Translationese classification accuracy for all settings (average and standard deviation over 5 runs). All of the models were trained/fine-tuned for the translationese classification task.}
    \label{tab:classification}
\end{table}

\begin{figure*}[!ht]
    \centering
    \includegraphics[width=0.8\textwidth]{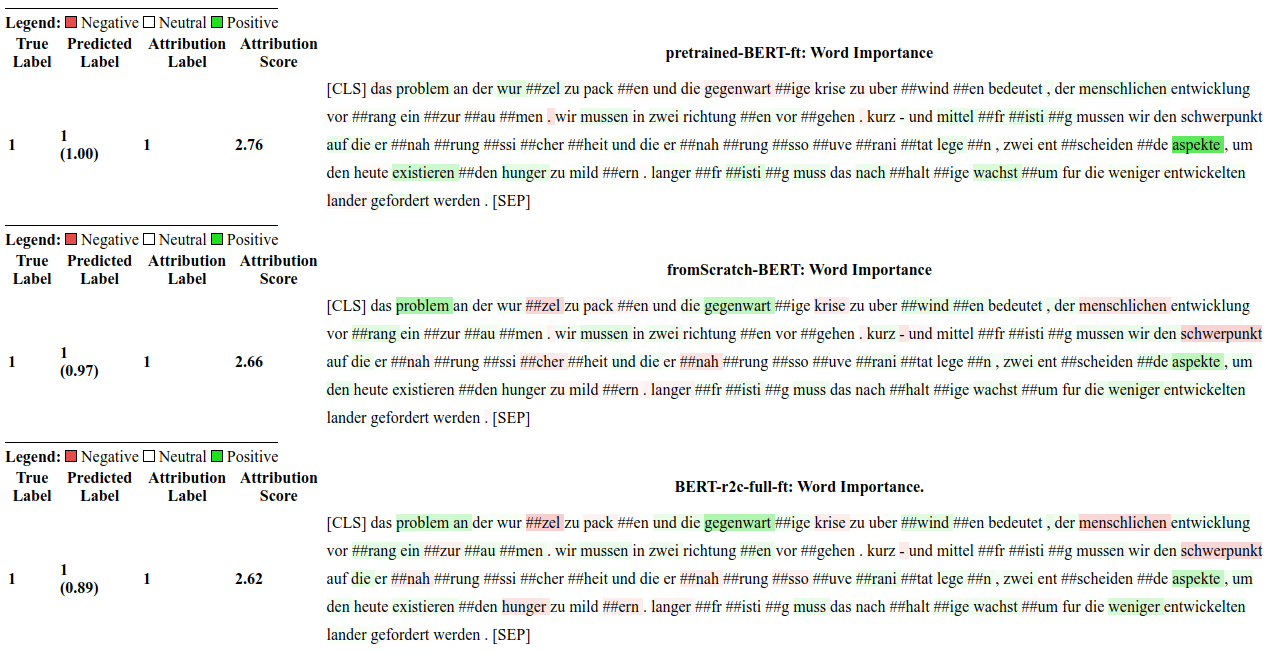}
    \includegraphics[width=0.8\textwidth]{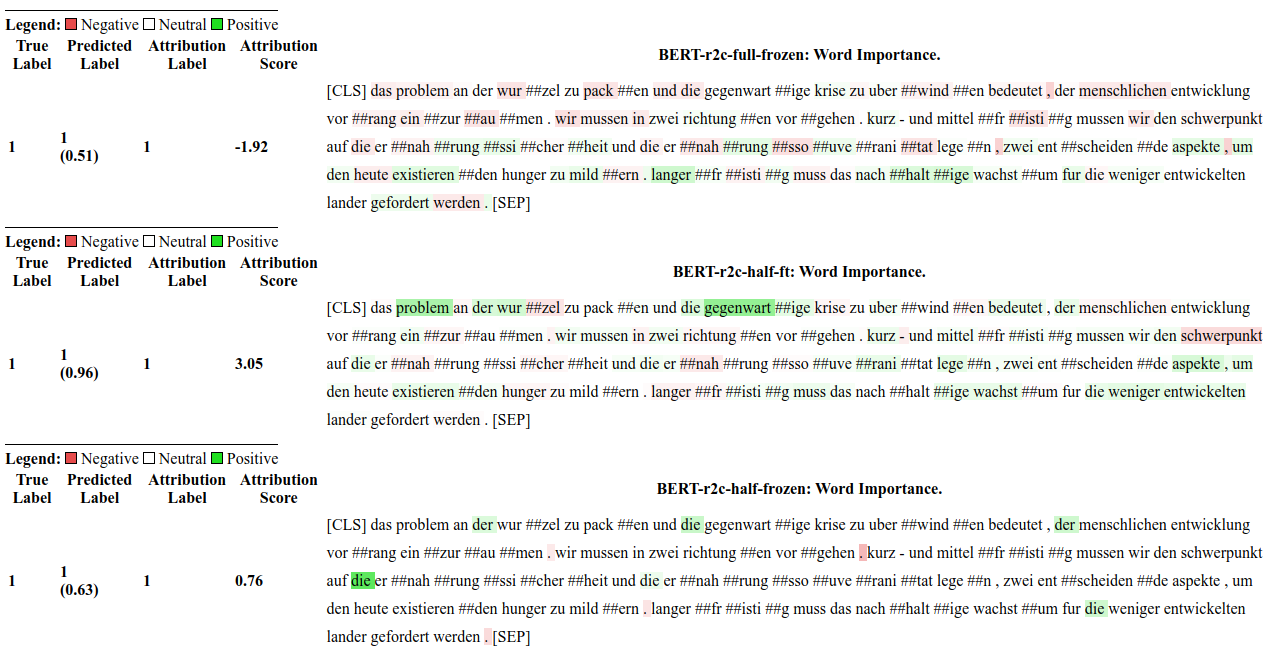}
    \caption{Layer Integrated Gradient saliency maps of input tokens contributing to the ground truth translationese label (here: translation). Comparison of different models.}
    \label{fig:attributions-models-trans}
\end{figure*}

\section{Translationese Classification}
\label{section:translationese_classification}

Table \ref{tab:classification} summarises results of the different translationese classification settings. For the base models, BERT outperforms the SVM by 16\% when trained from scratch and 19\% when finetuned.

Feeding pooled output of BERT into the SVM model [\textbf{pretrained-BERT-ft+SVM}], accuracy increases by 19\% percentage points absolute over using handcrafted features [\textbf{handcr.-features+SVM}], even when PCA is used to reduce the BERT vector dimensionality to match the size of the handcrafted feature vector. Feeding handcrafted features directly to the linear BERT classification head [\textbf{handcr.-features+LinearClassifier}] reduces accuracy by about 20\% points compared to pretrained and fine-tuned BERT [\textbf{pretrained-BERT-ft}]. This shows that features learnt by BERT are superior to our set of manual features, as used in previous high performing classical feature engineering-based approaches to translationese classification.
When BERT is trained from scratch on the MPDE data [\textbf{fromScratch-BERT}], translationese classification accuracy reduces by $\sim 3$ percentage points, compared to pretrained-BERT-ft. This suggests that pretraining on large data helps to encode additional information that turns out to be helpful in the translationese classification task.

One can assume that BERT pretrained to predict the handcrafted features and subsequently frozen [\textbf{BERT-r2c-full-frozen}] has learnt to encode the handcrafted features during pretraining (Section~\ref{ss:SVMintoBERT}). 
Nevertheless, its accuracy, albeit higher than a random guess, is lower by $\sim 13$ percentage points than the SVM classifier.
We perform an additional experiment, in order to check whether the difference in accuracy is due to BERT failing to sufficiently encode the handcrafted features during pretraining, or due to the SVM classifier being superior to the linear classification head of the BERT model. Namely, we train the SVM classifier on the pooled output of BERT-r2c-full-frozen model $[$\textbf{BERT-r2c-full-frozen+SVM}$]$ and on the PCA-reduced dimensionality $[$\textbf{BERT-r2c-full-frozen+PCA$_{108}$+SVM}$]$. The accuracy is around 75\% for both settings which is as high as using SVM on handcrafted feature vectors. We conclude that BERT encodes the handcrafted features sufficiently well, but the linear classifier performs worse than an SVM in these conditions.

Further fine-tuning BERT fully pretrained for handcrafted feature prediction [\textbf{BERT-r2c-full-ft}] for translationese classification results in accuracy comparable to BERT that was not pretrained on this task [\textbf{fromScratch-BERT}]. This could suggest that our handcrafted feature set is either a subset of features learned by fromScratch-BERT, or that the handcrafted features are discarded during fine-tuning. The model where only the first 6 layers were pretrained [\textbf{BERT-r2c-half-ft}], achieves similar accuracy, likely due to the same reasons.
By contrast, freezing the 6 handcrafted feature prediction pretrained layers [\textbf{BERT-r2c-half-frozen}] largely reduces the accuracy with respect to BERT-r2c-half-ft, because the model only has access to the 6th layer embeddings that supposedly encode the information about the handcrafted features, and does not have ability to extract its own features. The remaining (higher) 6 layers are responsible for the increment in accuracy with respect to BERT-r2c-full-frozen.

The results of BERT-f2c models show that BERT, when fed the handcrafted features in the form of embeddings, can reach at most the same accuracy as the hancdr.-features+SVM approach, which suggests that the BERT architecture has no advantage over the SVM classifier in utilizing the handcrafted features for classification. This is again evidence that the features, and not the classifier, cause the better performance of the feature and representation learning method.%
\footnote{As a sanity check, we ran an experiment using a gradient boosting classifier instead of an SVM, with the exact same 108 hand-crafted features and obtain accuracy of 72.3\%.}

\begin{table*}[ht]
    \centering
    \resizebox{0.80\textwidth}{!}{%
    \begin{tabular}{||l||c|c||c|c||c|c||c|c||}
    \hline
    \hline
    \multicolumn{1}{||c||}{} & \multicolumn{4}{c||}{\textbf{Translationese}} & \multicolumn{4}{c||}{\textbf{Original}} \\
    \hline
    \hline
    \multicolumn{1}{||c||}{} & \multicolumn{2}{c||}{\textbf{BERT-r2c-full-ft}} & \multicolumn{2}{c||}{\textbf{pretrained-BERT-ft}} & \multicolumn{2}{c||}{\textbf{BERT-r2c-full-ft}} & \multicolumn{2}{c||}{\textbf{pretrained-BERT-ft}} \\ \hline
        \textbf{Rank} & \textbf{Token} & \textbf{AAS} & \textbf{Token} & \textbf{AAS} & \textbf{Token} & \textbf{AAS} & \textbf{Token} & \textbf{AAS} \\ \hline
        1 & sagte & 0.60 & entstand & 0.70 & \#\#wegen & 0.61 & situations & 0.37 \\ \hline
        2 & gebiet & 0.46 & virus & 0.63 & • & 0.55 & • & 0.36 \\ \hline
        3 & \#\#dies & 0.44 & inti & 0.60 & eu & 0.49 & ria & 0.34 \\ \hline
        4 & ansicht & 0.43 & sagte & 0.58 & daraufhin & 0.49 & \#\#lk & 0.33  \\ \hline
        5 & bezug & 0.42 & entdeckte & 0.57 & finde & 0.45 & \#\#iet & 0.32 \\ \hline
        6 & neige & 0.40 & gras & 0.57 & \#\#vo & 0.45 & golden & 0.32 \\ \hline
        7 & amt & 0.40 & nuts & 0.56 & gerne & 0.43 & sak & 0.30 \\ \hline
        8 & pre & 0.40 & nicaragua & 0.55 & \#\#abb & 0.42 & turm & 0.30 \\ \hline
        9 & spanien & 0.39 & rekord & 0.53 & \#\#hrte & 0.42 & \#\#emen & 0.27 \\ \hline
        10 & sprechen & 0.38 & bilbao & 0.53 & ausbau & 0.42 & orange & 0.27 \\ \hline
        11 & nuts & 0.36 & verfugte & 0.53 & ! & 0.42 & hang & 0.26 \\ \hline
        12 & barcelona & 0.34 & bol & 0.51 & bekommen & 0.42 & \#\#wald & 0.25 \\ \hline
        13 & ; & 0.33 & colombia & 0.51 & trips & 0.41 & 1732 & 0.25 \\ \hline
        14 & \#\#bien & 0.32 & nis & 0.51 & ez & 0.41 & dobe & 0.24 \\ \hline
        15 & spanischen & 0.32 & och & 0.49 & \#\#gemeinde & 0.40 & \#\#pas & 0.23 \\ \hline
        16 & wiederholt & 0.31 & vorkommen & 0.49 & vot & 0.36 & profits & 0.22 \\ \hline
        17 & einige &  0.30 & oecd & 0.49 & won & 0.36 & stuttgart & 0.22 \\ \hline
        18 & \#\#sprache & 0.29 & ; & 0.46 & geplant & 0.35 & soja & 0.21 \\ \hline
        19 & weder & 0.29 & erklarte & 0.45 & demnach & 0.35 & r & 0.21 \\ \hline
        20 & territorium & 0.28 & clinton & 0.45 & ja & 0.35 & ruth & 0.21 \\ \hline \hline
    \end{tabular}
    }
    \caption{Top-20 tokens with highest average attribution score (AAS) towards original and translationese classes in the test set. BERT-r2c-full-ft and pretrained-BERT-ft.}
    \label{tab:top_words_pretrained-handcr_trans_org}
\end{table*}

\section{Layer Integrated Gradients Saliency}
\label{s:ligs}

We compare input attributions of the ground truth classification label amongst \textbf{pretrained-BERT-ft}, \textbf{fromScratch-BERT} and four different settings of the translationese classification models pretrained on the handcrafted feature prediction task: \textbf{BERT-r2c-full-ft}, \textbf{BERT-r2c-full-frozen}, \textbf{BERT-r2c-half-ft} and \textbf{BERT-r2c-half-frozen}.  
We use Layer Integrated Gradients from the Captum library \cite{kokhlikyan2020captum}, which computes the attribution for all the individual neurons in the embedding layer, and calculate the salience score for each token by averaging the attributions over the embedding dimension.

\paragraph{Comparing Models.}
Figure \ref{fig:attributions-models-trans} displays Integrated Gradients attributions for a translated paragraph across different BERT models. 
The trends for the original paragraph are similar to those that we observe for the translated paragraph, therefore attributions for the original paragraph are given in Appendix \ref{ssA:saliency}.

Comparing the attributions of classification labels to sample inputs amongst the various settings of BERT, we observe that attributions are similar for \textbf{fromScratch-BERT} and the fine-tuned models: \textbf{BERT-r2c-full-ft} and \textbf{BERT-r2c-half-ft}. This suggests that fine-tuning "dissolves" the pre-learned information about the hand-crafted features in the \textbf{r2c} models, no matter how much of the model was pre-trained.
By contrast, freezing the weights in \textbf{BERT-r2c-full-frozen} and \textbf{BERT-r2c-half-frozen} resulted in very different attributions compared to the \textbf{fromScratch-BERT}. Since these frozen models only utilize the information they have learnt about the handcrafted features, this shows that this information is not identical to the information that \textbf{fromScratch-BERT} learns for the translationese classification task. For \textbf{BERT-r2c-half-frozen} the attributions are more peaked than for other models, with only a few tokens receiving large scores, and most tokens having scores close to zero. Notably, \textbf{pretrained-BERT-ft} displays a pattern that is overall similar to BERT trained from scratch, but some attributions are reversed, and the peaks are on different tokens. This supports the observation that off-the-shelf BERT pretrained on a large amount of data encodes some useful additional information.

For\textbf{ BERT-r2c-full-frozen}, a substantial number of tokens with negative attributions have positive attributions in the model trained from scratch and also the fine-tuned models. However some attributions overlap, which suggests that \textbf{fromScratch-BERT} may be using something like the handcrafted features. We investigate this further by examining the fine-tuning checkpoints.

\paragraph{Comparing Checkpoints.}
We aim to study how \textbf{fromScratch-BERT} learns information about translationese classification over the epochs, and how this compares to the fine-tuning of \textbf{BERT-r2c-full-ft}, when the information about the hand-crafted features is gradually modified over the epochs turn into the final feature set used for translationese classification.
In Appendix \ref{ssA:saliency} we provide additional results on examining training checkpoints for \textbf{fromScratch-BERT} and \textbf{BERT-r2c-full-ft} for an original and a translated paragraph.

Results indicate that for \textbf{fromScratch-BERT} some attributions change into their opposite during training, whereas for \textbf{BERT-r2c-full-ft} the pattern appears to be already settled from the early checkpoints onwards, and does not change much over the course of fine-tuning. This supports the hypothesis that the handcrafted features are a subset of features learnt by \textbf{fromScratch-BERT}, and thus provide a useful initialization of weights for fine-tuning for translationese classification.

\paragraph{Highest Average Attribution.}
In order to make the interpretation less local, and to generalize the observations, we compute the top tokens with highest attribution on average across the test set. The results for each class for best-performing models (\textbf{pretrained-BERT-ft} and \textbf{BERT-r2c-full-ft}) are given in Table \ref{tab:top_words_pretrained-handcr_trans_org}.

For German translationese data translated from Spanish, some top tokens correspond to the geographical areas, where Spanish is spoken, e.g. "spanien", "barcelona", "spanischen" for BERT-r2c-full-ft; "nicaragua", "colombia", "bilbao" for pretrained-BERT-ft. (Moreover, in this example it appears that off-the-shelf pretrained-BERT-ft, pretrained on the Wikipedia data, better utilizes the non-European toponyms, unlike the BERT-r2c-full-ft that was only trained on the European-focused Europarl data.) Likewise for original German data, some of the top tokens are German geographical names, e.g. "stuttgart" for pretrained-BERT-ft. The subword "\#\#wald" also appears to be a common German toponymic suffix. This suggests that topic is one of the spurious clues that is used by BERT to determine the correct translationese class. This is also supported by the fact that some nouns that likely correspond to certain recurring discussion topics for only one class within our data sample, receive high attribution, e.g. "virus", "soja", "clinton", "orange" etc. The "ez" token, salient for the original class, appears to be a starting subword unit of the \textit{EZB} abbreviation (Europäische Zentralbank).

The "•" token (bullet point) having a high attribution for the class \textit{originals} for both models might suggest a spurious correlation within the dataset, that is apparently utilized by BERT. The ";" token is deemed important for the translationese class by both models, which might also be a spurious correlation. Conversely, this could be an indication that clauses in Spanish are more often joint with the semi-colon, than in German, which was preserved in the translation. 
This corroborates findings from other works that deep networks exploit spurious statistical cues for better performance \cite{mudrakarta-etal-2018-model, niven-kao-2019-probing}.

For both models the Präteritum forms "sagte", "erklärte" etc. are also among the top tokens important for recognizing translationese. One possible explanation could be that the Perfekt form ("hat gesagt") is more common in German spoken language, and Präteritum is more common in writing. Therefore the translators, while translating Spanish speeches into German, may have preferred to use the Präteritum form more common for writing.

\section{Summary and Conclusions}
We address two open questions in classification-based translationese research: (1) are the substantial performance differences between feature- and representation-learning and classical handcrafted feature based approaches is due to $(i)$ the difference in the features, $(ii)$ the classifiers, or $(iii)$ both, and (2) what do feature- and representation-learning based approaches actually learn? 

We address (1) by exchanging features from both models, examining a broad variety of settings, to ensure that this is done in a fair an unbiased way. We show that SVMs perform as good as BERT when fed with features learnt by BERT. Likewise, the BERT classification head and the full BERT architecture perform at the level of traditional SVM-based classification with handcrafted features, when fed with handcrafted features only. This shows that it is the feature and representation learning and not the classifiers that are responsible for the translationese classification performance difference.  

To address question (2), we examine BERT's input attributions using Integrated Gradients Saliency for various settings and observe that attributions are indeed similar for the model trained from scratch (fromScratch-BERT) on just the text data and the fine-tuned models that were pretrained on handcrafted feature prediction (BERT-r2c-full-ft and BERT-r2c-half-ft). This suggests that pretraining on the handcrafted features does not make a visible difference in attributions, and, together with the accuracy result that also does not change, suggests that no extra information is learnt during pretraining on handcrafted features. Based on these findings, and the fact that some attributions appear to overlap for BERT pretrained on handcrafted features and where the pretrained layers were subsequently frozen (BERT-r2c-full-frozen), and BERT trained from scratch (fromScratch-BERT), it is consistent to assume that handcrafted features are a (possibly partial) subset of the features automatically learnt by BERT.

Finally, analysis of top activated tokens suggests that at least part of BERT's strong translationese classification accuracy is based on topic differences between the classes as well as on some spurious correlations, rather than "proper" translationese phenomena. We are currently working on quantifying the 'Clever Hans' behaviour using named entity masking and cleaning/normalizing the data.

 \section*{Acknowledgements}

We would like to thank the reviewers for their insightful comments and feedback. This research is funded by the German Research Foundation (Deutsche Forschungsgemeinschaft) under grant SFB 1102: Information Density and Linguistic Encoding.

\bibliography{anthology,custom}
\bibliographystyle{acl_natbib}

\newpage
\appendix
\section{Appendix}
\label{sec:appendix}

\subsection{Extra Information on the MPDE Dataset}
\label{ssA:data}
We use version 2.0.0 of the \href{https://doi.org/10.5281/zenodo.5596238}{MPDE dataset} licensed under CC-BY 4.0. Specifically we use the \textit{mono\_de\_es} train/dev/test splits of the German-Spanish language pair. Table \ref{tab:mpde-stats} contains summary statistics of the data.

\begin{table}[!h]
    \centering
    \begin{tabular}{|l|c|}
    \hline
        Split & Number of Examples \\ \hline
        Train set & 29580 \\ \hline
        Validation set & 6366 \\ \hline
        Test & 6344 \\ \hline
    \end{tabular}
    \caption{Dataset statistics.}
    \label{tab:mpde-stats}
\end{table}

\subsection{Extra Information on BERT Models}
\label{ssA:bert}
With the exception of pretrained-BERT-ft, we use the \textit{transformers} library.%
\footnote{\url{https://huggingface.co/transformers/model_doc/bert.html}} Training is done across 4 NVIDIA GeForce GTX TITAN X GPUs with a batch size of 8 per GPU. We use a learning rate of $3 \cdot 10^{-5}$ and train or fine-tune for 5 epochs. Table \ref{tab:params} shows the number of parameters of the different BERT variants. Parameter counts include the embedding and respective prediction (classifier or regression) layers.

\begin{table}[!h]
    \centering
    \resizebox{0.45\textwidth}{!}{%
    \begin{tabular}{|l|l|c|c|}
    \hline
        Model & Num. Params (M) \\ \hline
        fromScratch-BERT & 177.85 \\ \hline
        BERT-reg-full & 177.94 \\ \hline
        BERT-reg-half & 135.41 \\ \hline
        BERT-r2c-* & 177.85 \\ \hline
        BERT-f2c $L = 1$ & 177.46 \\ \hline
        BERT-f2c $L = 80$ & 177.52 \\ \hline
        BERT-f2c $L = 256$ & 177.66 \\ \hline
        pretrained-BERT-f2c $L = 80$ & 177.52 \\ \hline
    \end{tabular}
    }
    \caption{Number of parameters of the various BERT models.}
    \label{tab:params}
\end{table}

\clearpage
\onecolumn
\subsection{Additional Layer Integrated Gradients saliency maps}
\label{ssA:saliency}

\bigskip
\begin{figure*}[!h]
    \centering
    \includegraphics[width=0.9\textwidth]{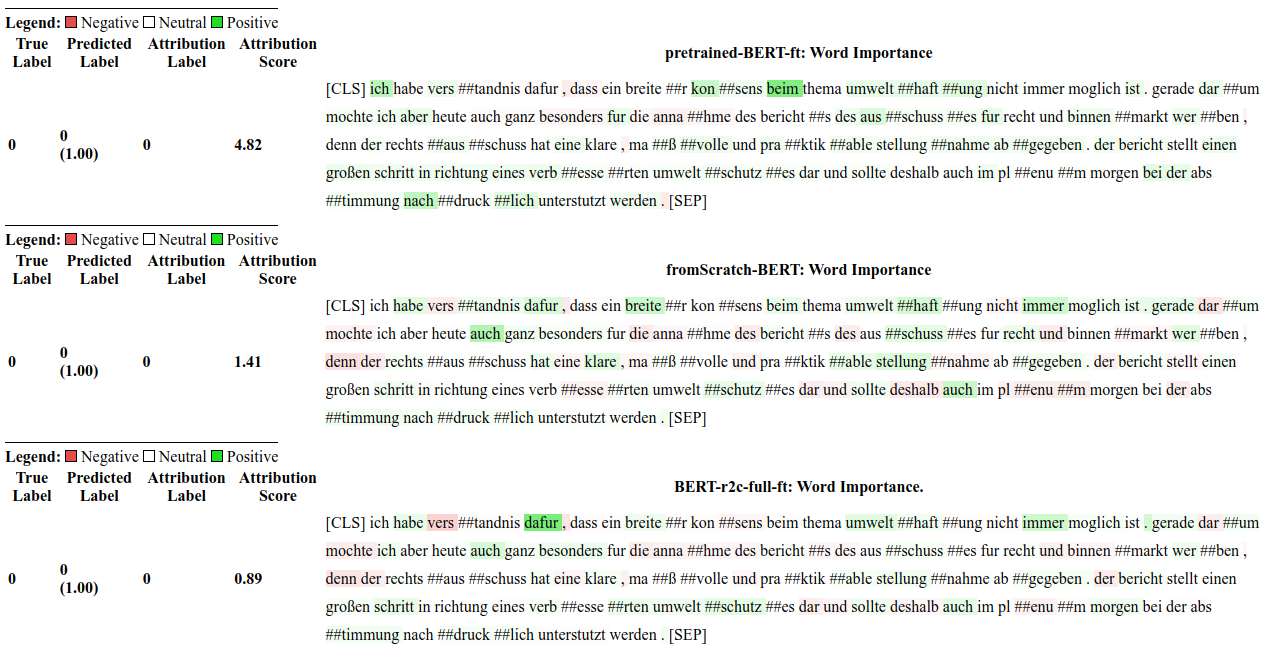}
    \includegraphics[width=0.9\textwidth]{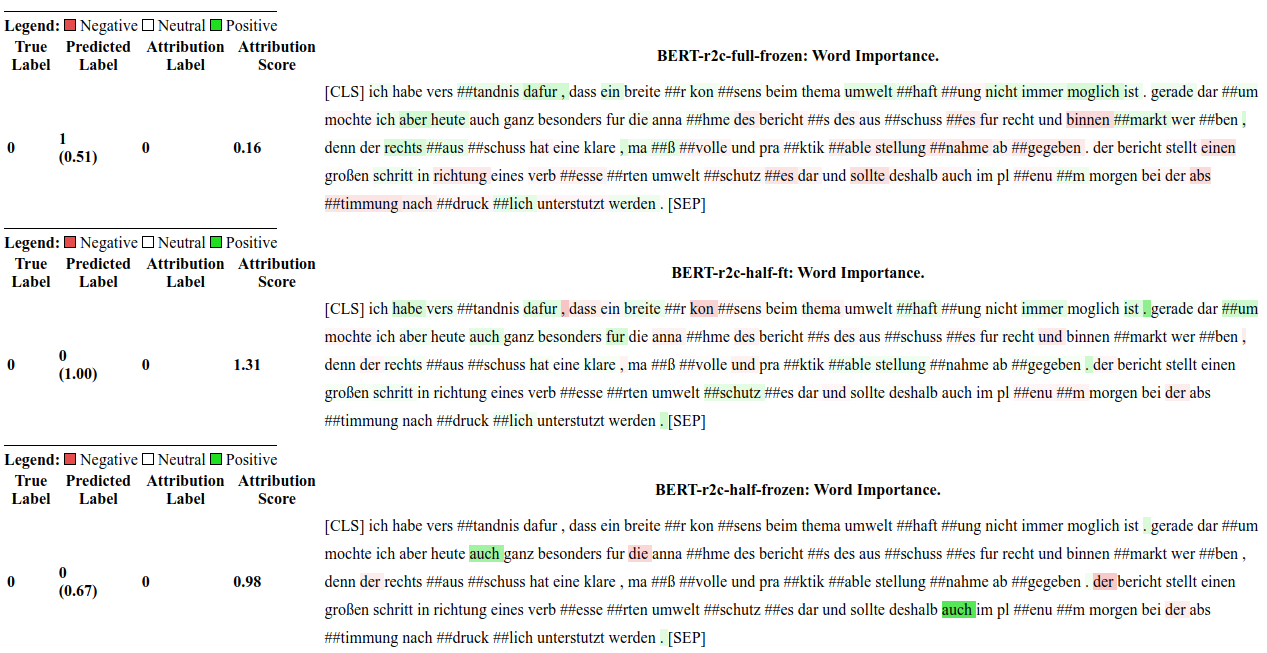}
    \caption{Layer Integrated Gradient saliency maps of input tokens contributing to the ground truth translationese label (here: original). Comparison of different models.}
    \label{fig:attributions-models-org}
\end{figure*}

\clearpage
\begin{figure*}[!h]
    \centering
    \includegraphics[width=0.9\textwidth]{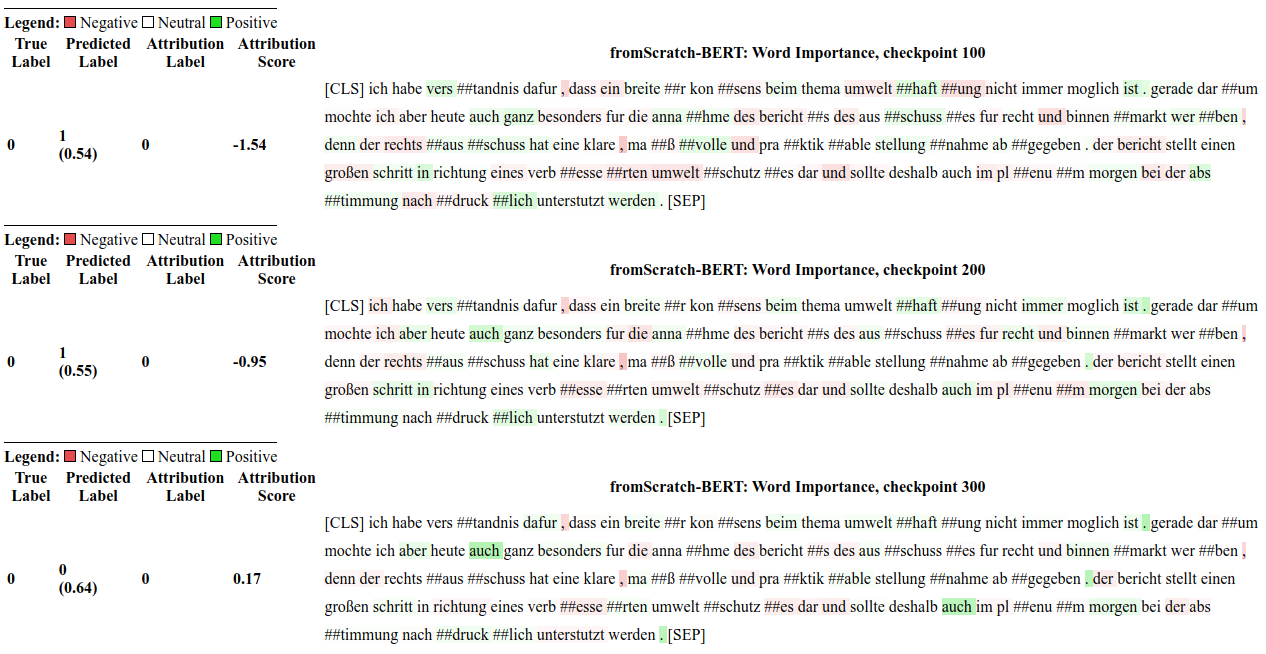}
    \includegraphics[width=0.9\textwidth]{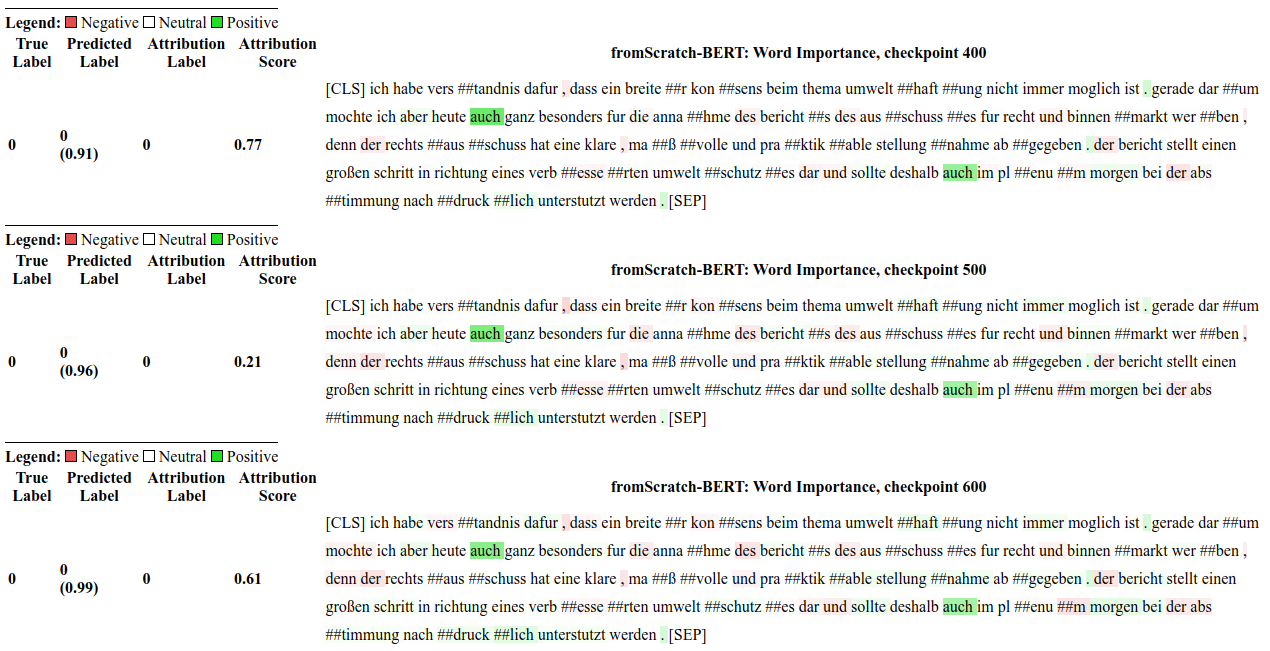}
    \includegraphics[width=0.9\textwidth]{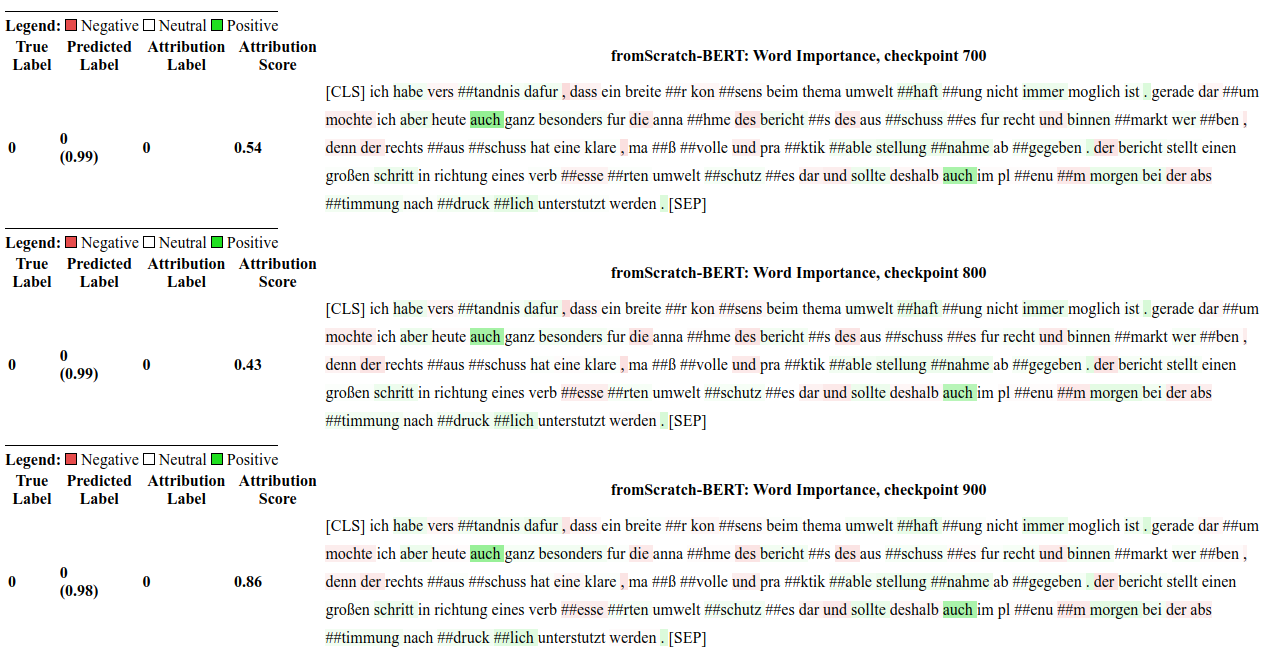}
    \caption{Layer Integrated Gradient saliency maps of input tokens contributing to the ground truth translationese label (here: original). BERT trained from scratch for translationese classification. Changes in attribution over the training checkpoints.}
    \label{fig:attributions-checkpoints-scratch-org}
\end{figure*}

\clearpage
\begin{figure*}[!h]
    \centering
    \includegraphics[width=0.9\textwidth]{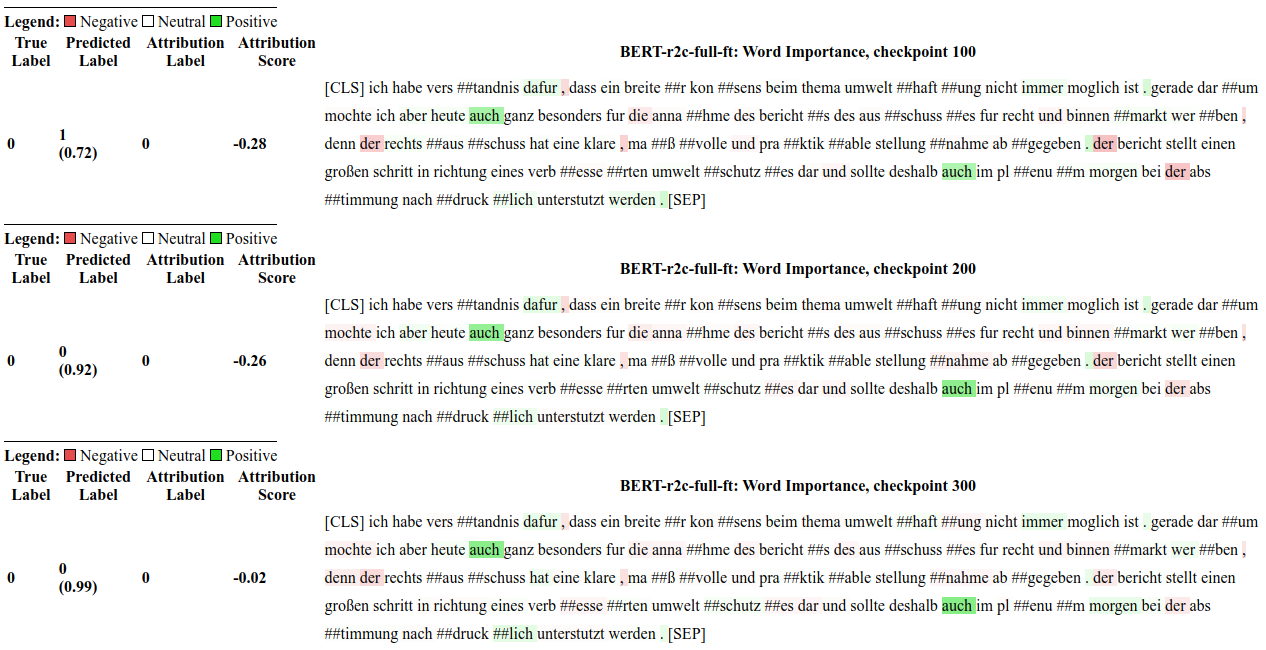}
    \includegraphics[width=0.9\textwidth]{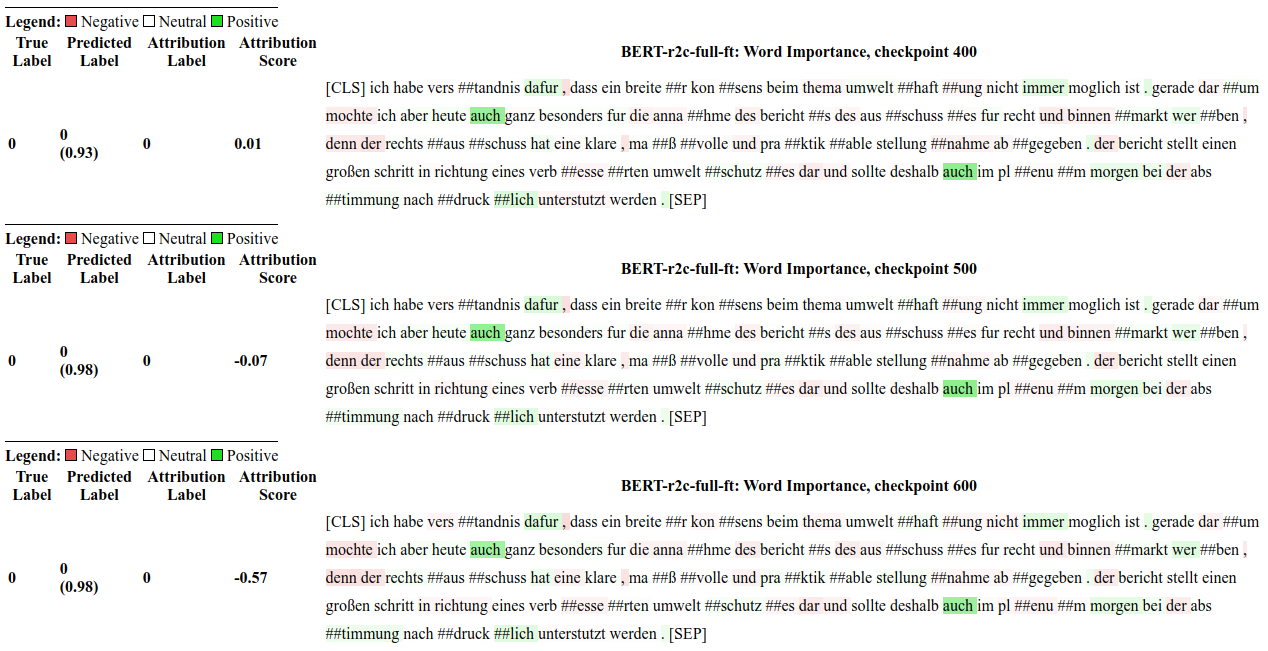}
    \includegraphics[width=0.9\textwidth]{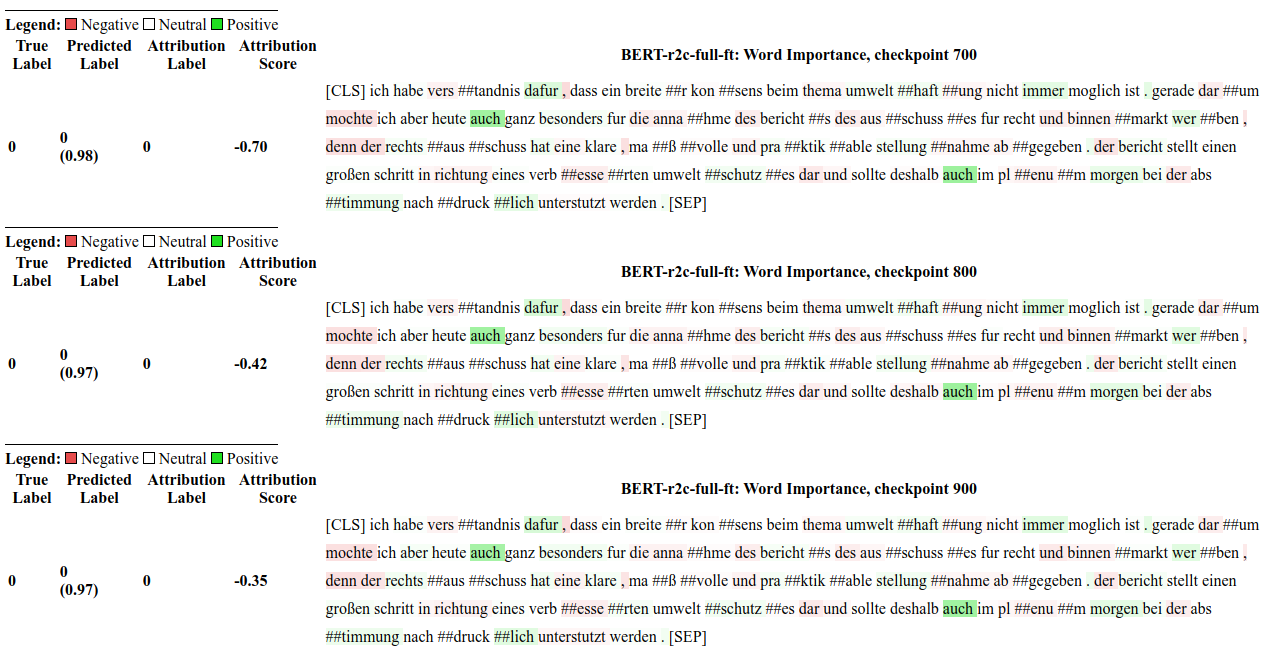}
    \caption{Layer Integrated Gradient saliency maps of input tokens contributing to the ground truth translationese label (here: original). BERT pretrained for handcrafted feature prediction, and fine-tuned for translationese classification. Changes in attribution over the training checkpoints.}
    \label{fig:attributions-checkpoints-pretrained-org}
\end{figure*}

\clearpage
\begin{figure*}[!h]
    \centering
    \includegraphics[width=0.9\textwidth]{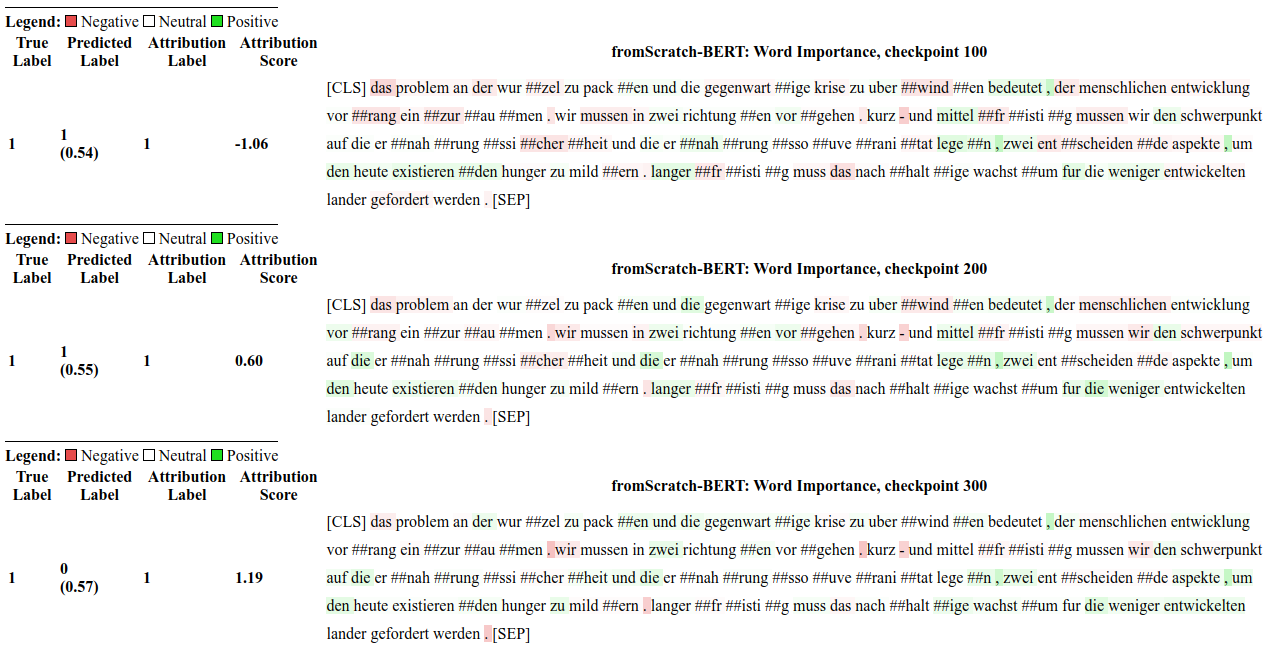}
    \includegraphics[width=0.9\textwidth]{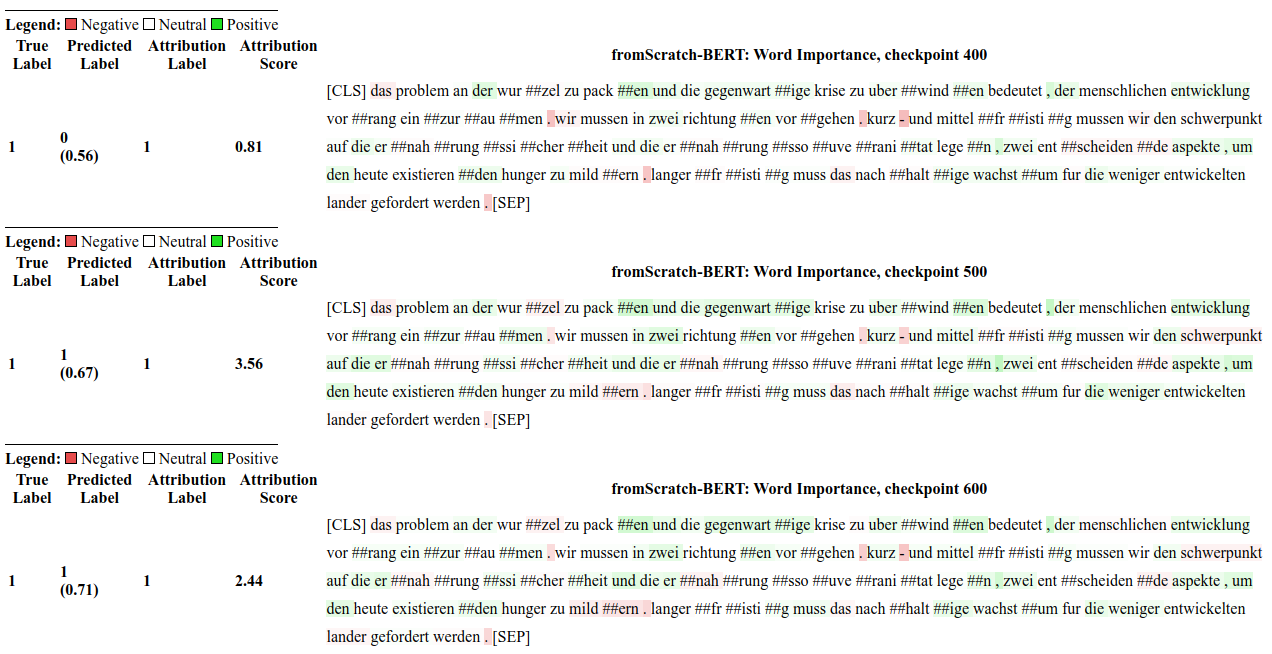}
    \includegraphics[width=0.9\textwidth]{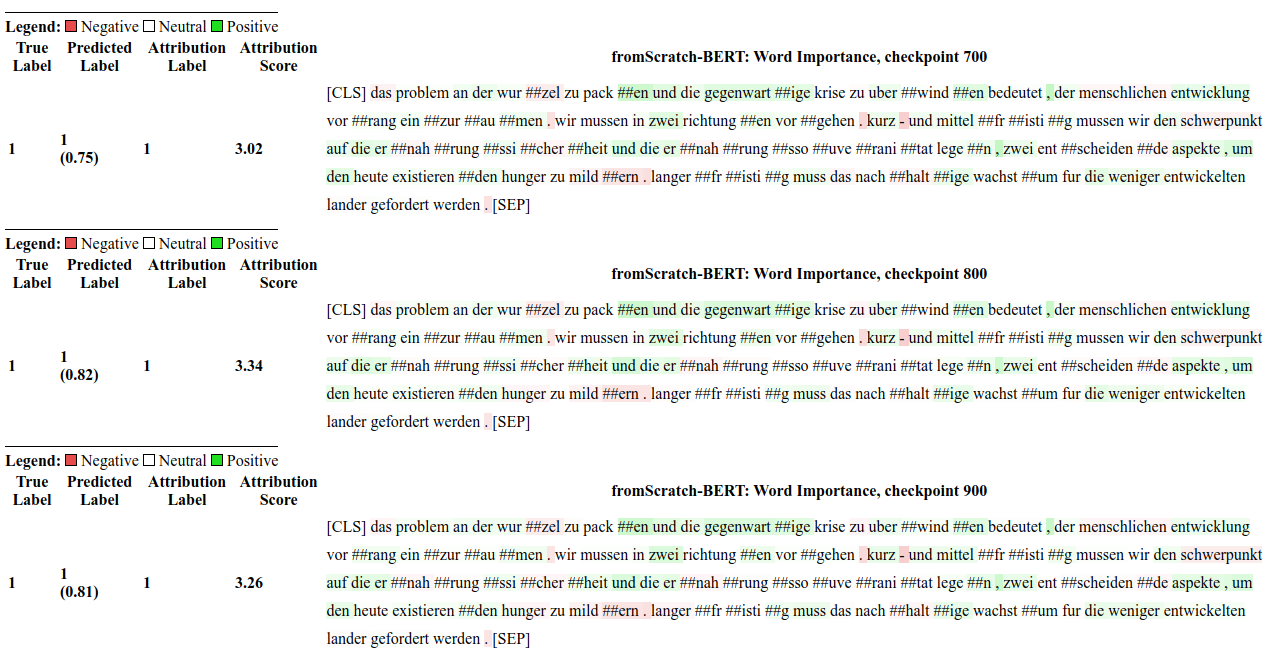}
    \caption{Layer Integrated Gradient saliency maps of input tokens contributing to the ground truth translationese label (here: translation). BERT trained from scratch for translationese classification. Changes in attribution over the training checkpoints.}
    \label{fig:attributions-checkpoints-scratch-trans}
\end{figure*}

\clearpage
\begin{figure*}[!h]
    \centering
    \includegraphics[width=0.9\textwidth]{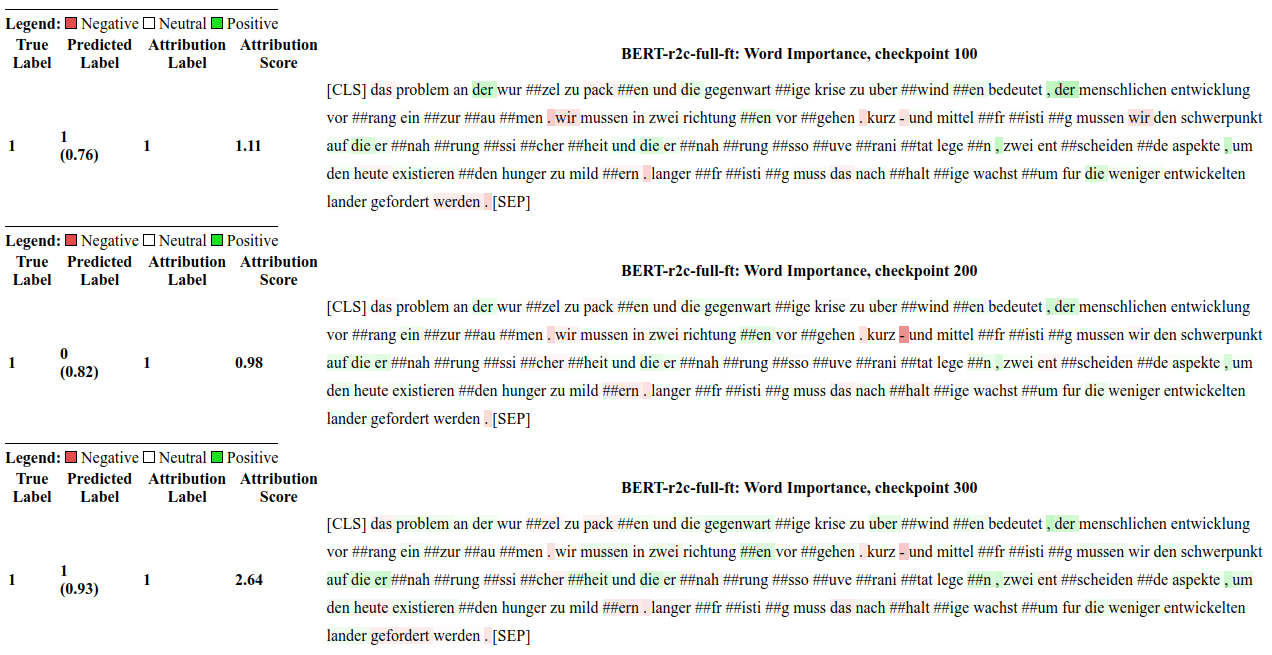}
    \includegraphics[width=0.9\textwidth]{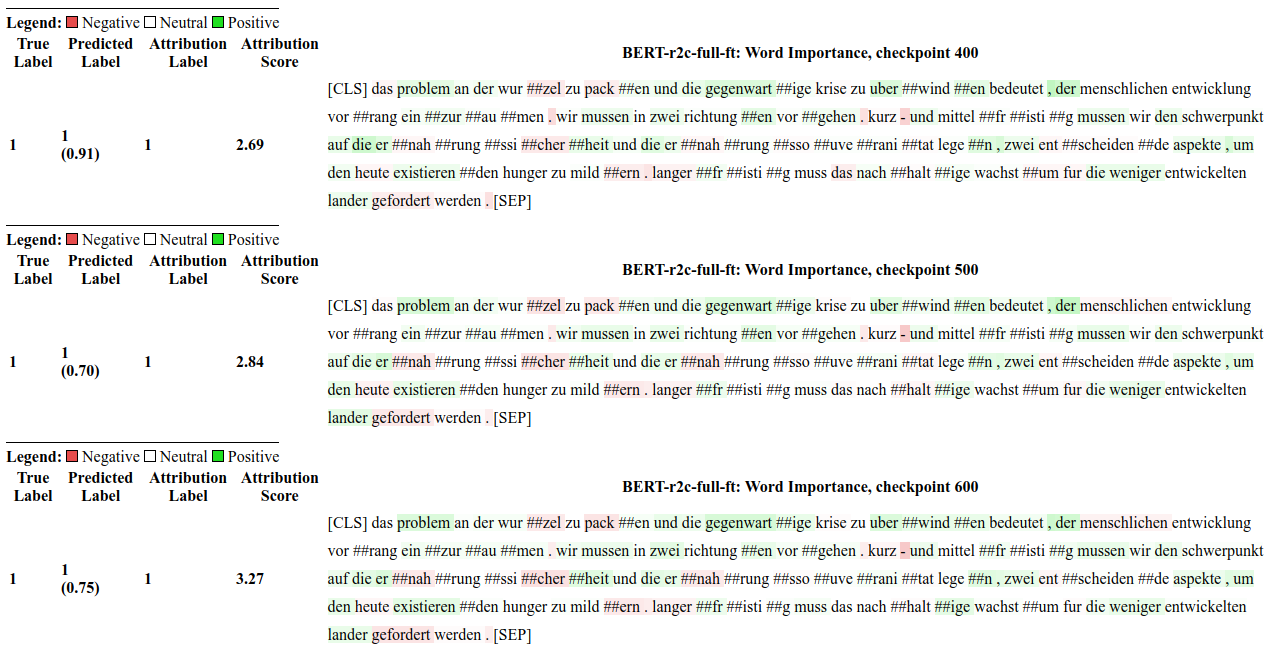}
    \includegraphics[width=0.9\textwidth]{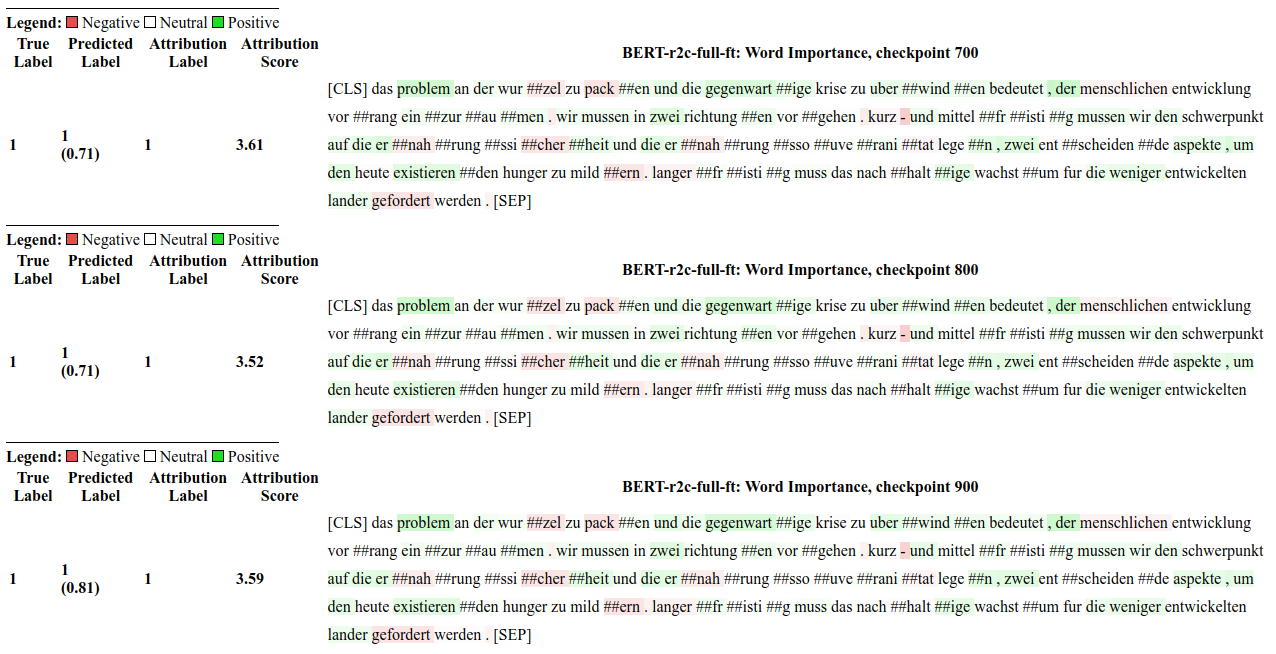}
    \caption{Layer Integrated Gradient saliency maps of input tokens contributing to the ground truth translationese label (here: translation). BERT pretrained for handcrafted feature prediction, and fine-tuned for translationese classification. Changes in attribution over the training checkpoints.}
    \label{fig:attributions-checkpoints-pretrained-trans}
\end{figure*}

\end{document}